\documentclass[12pt,a4paper]{article}

\usepackage[british]{babel}

\usepackage[a4paper,top=2cm,bottom=2cm,left=2.5cm,right=2.5cm,marginparwidth=1.75cm]{geometry}

\usepackage[citestyle=numeric, bibstyle=ama]{biblatex} % AMA 10th edition style citations using biblatex
\DeclareLanguageMapping{british}{british-apa} % Set language mapping
\usepackage{listings}
\usepackage{amsmath}
\usepackage{graphicx}
\usepackage[colorlinks=true, allcolors=blue]{hyperref}
\usepackage{hyperref}
\usepackage[title]{appendix}
\usepackage{mathrsfs}
\usepackage{amsfonts}
\usepackage{booktabs} % For \toprule, \midrule, \botrule
\usepackage{caption}  % For \caption
\usepackage{threeparttable} % For table footnotes
\usepackage{algorithm}
\usepackage{algorithmicx}
\usepackage{algpseudocode}
\usepackage{listings}
\usepackage{enumitem}
\usepackage{chngcntr}
\usepackage{booktabs}
\usepackage{lipsum}
\usepackage{subcaption}
\usepackage{authblk}
\usepackage[T1]{fontenc}    % Font encoding
\usepackage{csquotes}       % Include csquotes
\usepackage{diagbox}
\usepackage{color,soul}
\usepackage{svg}
\usepackage{comment}

\usepackage{siunitx}

\usepackage{setspace}
\usepackage{titlesec}
\titleformat{\section} % Redefine section numbering format
  {\normalfont\Large\bfseries}{\thesection.}{1em}{}
  
\usepackage{lineno} % Add the lineno package

\rightlinenumbers % Right-align line numbers

\newcommand{\subsubsubsection}[1]{%
  \vspace{\baselineskip}% Add some space
  \noindent\textbf{#1\\}\quad% Adjust formatting as needed
}
\usepackage{float}   % for customizing caption position
\usepackage{caption} % for customizing caption format
\usepackage{multirow}
\usepackage{longtable}
\usepackage{graphicx}



\title{Lettuce: An Open Source Natural Language Processing Tool for the Translation of Medical Terms into Uniform Clinical Encoding}

\author[1,2,3]{James Mitchell-White}
\author[2,3]{Reza Omidvar}
\author[1,3]{Benjamin Partridge}
\author[1,3]{Esmond Urwin}
\author[2]{Karthikeyan Sivakumar}
\author[3,4]{Ruizhe Li}
\author[1]{Andy Rae}
\author[5]{Xiaoyan Wang}
\author[5]{Theresia Mina}
\author[1,3]{Tom Giles}
\author[7]{Diego García-Gil}
\author[1]{Tim Beck}
\author[5,6]{John Chambers}
\author[1,3.*]{Grazziela Figueredo}
\author[1,3]{Philip R Quinlan}
\affil[1]{\small Centre for Health Informatics, School of Medicine, The University of Nottingham}
\affil[2]{Digital Research Service, The University of Nottingham}
\affil[3]{NIHR Nottingham Biomedical Research Centre}
\affil[4]{School of Computer Science, The University of Nottingham}
\affil[5]{Lee Kong Chian School of Medicine, Nanyang Technological University, Singapore}
\affil[6]{Department of Epidemiology and Biostatistics, School of Public Health, Imperial College London, United Kingdom}
\affil[7]{Department of Software Engineering, Andalusian Institute of Data Science and Computational Intelligence (DaSCI), University of Granada, 18012, Granada, Spain}
\affil[*]{Corresponding author: \texttt{g.figueredo@nottingham.ac.uk}}

\begin{document}
\maketitle

\begin{abstract}
\textbf{Background:} Existing solutions for finding the standardized vocabulary concepts for OMOP, such as the Athena database search and Usagi struggle with semantic nuances and require substantial manual input.

\textbf{Objectives:} Lettuce was developed to be an open-source tool for mapping source terms to OMOP concepts with configurable pipelines using advanced natural language processing, including large language models and text search, to automate and enhance the mapping process.

\textbf{Methods:} The performance of lettuce pipelines was compared with representative lexical search methods using two drug source-term datasets.

\textbf{Results:} Semantic search pipelines consistently outperformed existing methods, with up to a two-fold increase in the correct concept being in the top 10 results.

\textbf{Conclusions:} This paper introduces Lettuce, an intelligent open-source tool designed to address the complexities of converting medical terms into OMOP standard concepts.
Developed with a focus on GDPR compliance, Lettuce can be deployed locally, ensuring data protection while maintaining high performance in converting informal medical terms to standardised concepts.

\end{abstract}


%-------------------------------------------
% Paper Body
%-------------------------------------------

%--- Section ---%
\section{Introduction}

Making data findable, accessible, interoperable, and reusable (FAIR)\parencite{Wilkinson2016FAIR} is best practice for data stewardship, and enables transparent, reproducible research. 
The Observable Medical Outcomes Partnership (OMOP) common data model (CDM) is an open community standard for making observational healthcare data FAIR~\parencite{OHDSI2024Data, jeffersonHybridArchitectureCOCONNECT2022}.
The process of converting data to OMOP, however, is complex, and not only requires knowledge of the specific domain of the data, but often collaboration from data engineers, software engineers, and healthcare professionals.
Unified data standards are often applied inconsistently across healthcare systems \parencite{Hardy2022Data,Cholan2022Encoding}, and different mapping choices contributes to this inconsistency.

In previous work, we developed Carrot Mapper and Carrot-CDM \parencite{coxConversionSensitiveData2025} to support the OMOP conversion process.
Tooling within this space still generally requires manual intervention to approve or create mappings, where a data engineer needs to find the most suitable codification for a medical term taken from the metadata of a source dataset, or "source term", though automated mapping tools are available~\parencite{adamsBreakingDigitalHealth2025,ahmedAutomatedOMOPConcept,Hyve}.
For transformation to OMOP, the mapping must be from this source term to one or more terms taken from the Observational Health Data Sciences and Informatics (OHDSI) standardised vocabularies, a set of medical vocabularies designed to represent all OMOP domains.
Solutions that help find codifications include searches in  the OHDSI Athena database~\parencite{athena_ohdsi}, or string matching using tools like Usagi~\parencite{usagi_documentation}. 

The Athena website is a platform for searching and exploring various medical terminologies, vocabularies, and concepts in healthcare research.
Users can search for specific terms, view their relationships, and explore detailed metadata.
Using Athena search at scale, however, is complicated.
When conducting extensive searches, researchers face challenges, including the complexity and overlap of medical vocabularies, the overwhelming volume of search results, and technical constraints, such as system performance and data handling capabilities.
Additionally, the standardisation of diverse healthcare datasets presents difficulties in ensuring consistency across different terminologies. 

Usagi was developed by OHDSI to facilitate the mapping of source terms to standard concepts within the OMOP CDM.
It supports the integration and harmonisation of diverse healthcare data sources.
Usagi employs semi-automated string-matching algorithms to suggest potential mappings between local vocabularies and standardised terminologies such as SNOMED CT, LOINC, and RxNorm.
It is a valuable tool for mapping, but it has a few limitations. 
While it automates part of the mapping process, it requires significant manual review, which is time-consuming and prone to human error and uncertainties.
String-matching can potentially lead to inaccurate mappings, particularly when dealing with ambiguous or complex terminologies.
The effectiveness of Usagi depends on the quality of the standardised vocabularies it uses, and there is a learning curve for new users.

Usagi works well when dealing with data with typographical errors, but not all source terms are in formalised language as used in OHDSI vocabularies, and these informal terms for medications or conditions may not closely match the string of the formal concept we wish to map it to.
For example, “Now Foods omega-3” is a supplement found in a self-reported patient questionnaire dataset.
This supplement is produced by Now Foods, and is an omega-3 product derived from fish oil.
In this case, the brand of the drug was given as input.
Before obtaining the OMOP concept, we need to map the reported brand to “omega-3 fatty acids”, for which an exact OMOP match is found.
Using the Athena search engine, for example, the string matching suggests concepts like “Calcium ascorbate 550 MG Oral Tablet by Now Foods”, “Ubiquinone 90 MG Oral Capsule [Now Coq10] by Now Foods” or “Calcium ascorbate 1000 MG Oral Tablet [Now Ester-C] by Now Foods”.
This indicates that this process of matching loses the semantic information associated with the input data. 

Large Language Models (LLMs) are a relatively novel alternative to support OMOP.
They automate portions of the mapping process while suggesting more semantically relevant mappings.
The use of proprietary tools, such as OpenAI ChatGPT~\parencite{chatgpt} in healthcare, however, raises significant concerns, particularly regarding GDPR compliance, data protection and reproducibility of results~\parencite{Nazi2024LLMHealthcare,Deng2024LLMStatus}.
The handling of sensitive patient data poses risks, as inadvertent data leaks or misuse of information could occur.
Ensuring that interactions with OpenAI and other available LLM APIs in the cloud remain within the bounds of GDPR is challenging, especially when dealing with identifiable health information. 

In this paper we introduce Lettuce\footnote{https://github.com/Health-Informatics-UoN/lettuce}, a tool created to address these gaps.
It is a stand-alone, open-source, adaptable natural language processing tool based on LLMs, querying systems and text search for the conversion of medical terms into the OMOP standard vocabulary.
Lettuce is released under the MIT Licence.
Medical terms can be extracted from Electronic Health Records (EHRs), self-reported patient questionnaires and other structured datasets to serve as an input for Lettuce.
So for the example above, the Lettuce match output for “Now Foods omega-3” is “Fish oil”.

Lettuce has the following modules and functionalities:

\begin{itemize}
    \item Vector search for concept(s)
    \item LLM prompting with informal name(s)
    \item OMOP CDM database search
\end{itemize}

We demonstrate how Lettuce works and its performance compared to Usagi and ChatGPT on a case study of converting self-reported informal medication names into OMOP concepts. Lettuce performance is comparable to OpenAI models and was developed to run locally to support healthcare data governance requirements.

\section{Methods}
\label{sec:methods}
\subsection{Lettuce Architecture}

\begin{figure}[hb]
    \centering
    \includegraphics[width=1\linewidth]{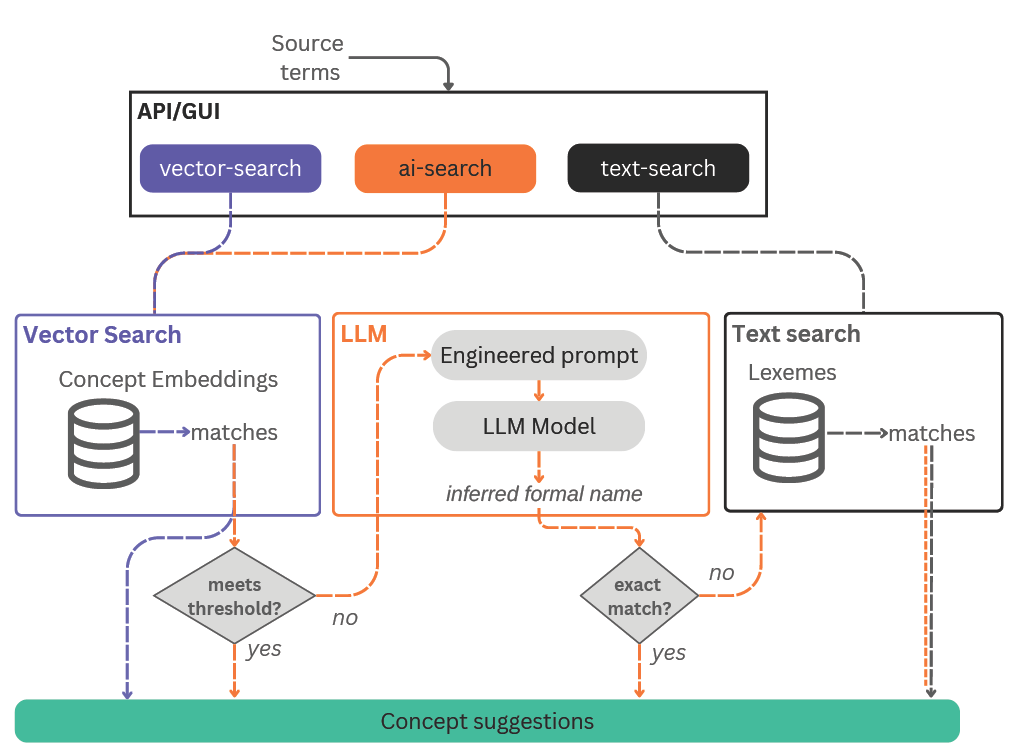}
    \caption{Lettuce concept mapping search routes}
    \label{fig:lettuce-architecture}
\end{figure}

The Lettuce architecture comprises an OMOP-CDM database and three routes for concept retrieval (Figure~\ref{fig:lettuce-architecture}). For each input term, the user selects one pathway to generate a ranked set of mapping recommendations.
The pathways are: (i) text search, (ii) vector-based semantic retrieval, and (iii) retrieval-augmented generation (RAG).

The system receives a source term—the free-text entity requiring mapping to an OMOP concept—together with optional constraints such as target domain or vocabulary.
The output is a ranked list of candidate concepts and their confidence score derived from the retrieval strategy employed.

\subsubsection{Database}

Lettuce operates on standardised vocabularies from the OMOP-CDM, which can be obtained from the OHDSI Athena repository~\parencite{athena_ohdsi}.
Each concept is stored as a row in the concept table, containing a unique identifier, a concept name, the source vocabulary, the domain of the concept, and additional metadata.

To support efficient retrieval, Lettuce extends the native OMOP schema with two additional components:

\begin{enumerate}
    \item Pre-computed lexical features, enabling accelerated keyword search.

    \item Pre-computed dense vector embeddings, implemented using PGVector~\parencite{PgvectorPgvector2026} to support similarity search.
\end{enumerate}

An OMOP-CDM database configured with these features can be deployed using omop-lite~\parencite{HealthInformaticsUoNOmoplite2025}.

\subsubsection{Text search}

This retrieval pathway uses keyword matching. Lettuce breaks the input term into its lexemes --- basic word units --- after reducing each word to its root form so that variations like ``fracture'' and ``fractured'' are treated the same.
The input is compared with the stored lexemes for each concept in the database. Concepts that share the most lexemes with the input are ranked highest, and the top-k matches are returned. Lettuce’s lexical search mechanism conducts keyword-based retrieval over concept names, optionally filtered by domain, vocabulary, or other attributes.

Storing pre-computed lexemes is much faster than computing them for every query, so a column of lexemes is materialised during database set-up.
This component works well when the input term uses similar wording to an existing OMOP concept. However, it is less effective when the correct concept uses different terminology—for example, when two terms mean the same thing but do not share similar wording.

\subsubsection{Vector-Based Semantic Search}

The second retrieval pathway identifies concepts based on semantic similarity rather than shared wording.
It uses deep learning models known as bidirectional encoders, which are trained to understand how words relate to one another within context.
These models convert text into numerical vectors, called embeddings.
In this high-dimensional space, terms with similar meanings are located close together, even when their written forms differ significantly~\parencite{bengioNeuralProbabilisticLanguage2003, devlinBERTPretrainingDeep2019}.

For example, paracetamol and acetaminophen are completely different strings, yet they refer to the same substance; their corresponding embeddings therefore should be close together. Table~\ref{tab:similaritydemo} shows the cosine similarity calculation between these terms, after they are embedded using BGE-small-en-v1.5~\parencite{bge_embedding}. By contrast, paracetamol and piracetam share some surface-level similarity in spelling, but have unrelated clinical meanings, and their embeddings are far apart as shown in the table's calculations.  

Computing embeddings is computationally expensive, so Lettuce generates embeddings for all the concepts in the CDM in advance, otherwise queries would be prohibitively slow.
These pre-computed vectors are stored in an embeddings table within the OMOP-CDM database. Each concept is represented by a string describing the concept, which is encoded using a sentence-transformer model~\parencite{reimersSentenceBERTSentenceEmbeddings2019}. When a user provides a source term, Lettuce encodes it using the same model and calculates its cosine similarity to the stored embeddings. The top-(k) closest matches are then returned, together with metadata from the concept table. Because the embeddings table is fully integrated into the database, vector search results can be filtered by domain or vocabulary in the same way as keyword search. Although faster approximate nearest-neighbour search algorithms exist, they introduce small inaccuracies and require substantial indexing time and storage\cite{malkovEfficientRobustApproximate2020}. Given that accuracy is critical for concept mapping, Lettuce opts for exact similarity search despite the higher computational cost.

\begin{table}
    \centering
    \begin{tabular}{l *{4}{S[table-format=1.2]}} 
    \toprule
        & {paracetamol} & {acetaminophen} & {piracetam} & {ibuprofen} \\
    \midrule
    paracetamol   & 1.00 & 0.83 & 0.70 & 0.74 \\
    acetaminophen &      & 1.00 & 0.73 & 0.76 \\
    piracetam     &      &      & 1.00 & 0.64 \\
    ibuprofen     &      &      &      & 1.00 \\
    \bottomrule
    \end{tabular}
    \caption{Example of cosine similarity values between pairs of words, encoded using BGE-small-en-v1.5~\parencite{bge_embedding}. Cosine similarity is symmetrical, so the bottom of this table is omitted.}
    \label{tab:similaritydemo}
\end{table}

\subsubsection{Large Language Model Search and Retrieval}

The final retrieval pathway combines semantic search with LLMs to handle cases where neither keyword nor embedding-based methods produce a high confidence match. LLMs are powerful general-purpose language tools, but they do not contain the OMOP vocabulary in their internal memory. To make them effective for concept mapping, Lettuce uses RAG~\cite{lewis2020rag}, a technique that supplies the model with relevant information at query time.

The process begins with vector search. If the top embedding is highly similar to the user’s input---above a configurable threshold---it is treated as a likely exact match and returned. If the similarity is lower, Lettuce selects the top semantically relevant concepts and inserts them into a prompt given to the LLM; in addition, the prompt asks for alternative solutions that are not in the list but could be suitable. These retrieved concepts act as context, guiding the model towards accurate, domain-specific reasoning.

The LLM then produces its suggestion for the appropriate concept. Lettuce validates this output by first checking whether it exactly matches an OMOP concept name. If it does, that concept is returned. If not, the system performs a keyword search on the model’s output and provides the top-k results along with the alternative results produced by the LLM. This ensures the model’s reasoning is always grounded in the actual OMOP vocabulary. Examples of the prompts used are shown in Section \ref{section:RAG}.

\subsection{Command-line interface (CLI)}
A command line interface can be used to run Lettuce, specifying arguments including which components to engage and which filters to apply to search results.
The only required argument is an \lstinline{--informal_name} to use as a search term.
For example:
\begin{lstlisting}
    uv run lettuce-cli --informal_name "Nasonex (for each nostril)"
\end{lstlisting}
will run an LLM-enabled search for ``Nasonex (for each nostril)''.

\subsection{HTTP Application Programming interface}
A Lettuce server can be started instead of the CLI, which then takes HTTP GET requests.
This API is implemented using FastAPI~\parencite{fastapi}.
The Lettuce server endpoints have a common format for input and output, designed for presentation by a graphical user interface.
The parameters for starting a search are described in Table \ref{tab:api_params}.
A lightweight, low-latency endpoint runs only the text-search feature (\lstinline{GET /search/text-search/<source-term>}), and is suitable for source terms that are lexically close to an OMOP concept.
A second endpoint runs the vector search (\lstinline{GET /search/vector-search/<source-term>}), which can be useful for suggesting concepts that are semantically similar to a source term, even if they are lexically dissimilar.
The final endpoint (\lstinline{GET /search/ai-search/<source-term>} runs the RAG pipeline, followed by text search on the LLM output

The format of the output is a list of suggested concepts, which optionally includes the scores and ranks from the functions used to rank suggestions, and metadata describing the application, version, and optionally additional information (described in Tables \ref{tab:concept_fields}, \ref{tab:metadata_fields} and \ref{tab:suggestions_structure}).
In the RAG-based endpoint, the additional information is used to communicate the LLM's response.

\begin{table}[h]
    \centering
    \begin{tabular}{llp{8cm}}
    \toprule
        \textbf{Field} & \textbf{Type} & \textbf{Description} \\
    \midrule
        search\_term & str & The source term to search for  \\
    \midrule
        vocabulary & List[str] | None & A list of vocabularies to include. If None, includes all vocabularies \\
    \midrule
        domain & List[str] | None & A list of domains to include. If None, includes all domains \\
    \midrule
        standard\_concept & bool & If true, only includes standard concepts \\
    \midrule
        valid\_concept & bool & If true, excludes concepts that have been invalidated \\
    \midrule
        top\_k & int & The number of concepts to fetch \\
    \bottomrule
    \end{tabular}
    \caption{Parameters of a request to Lettuce's API}
    \label{tab:api_params}
\end{table}

\begin{table}[ht]
\centering
\begin{tabular}{l l p{8cm}}
\toprule
\textbf{Field} & \textbf{Type} & \textbf{Description} \\
\midrule
concept\_name & str & The concept\_name field from the OMOP concept table \\
\midrule
concept\_id & int & The concept\_id field from the OMOP concept table \\
\midrule
domain\_id & str & The domain\_id field from the OMOP concept table \\
\midrule
vocabulary\_id & str & The vocabulary\_id field from the OMOP concept table \\
\midrule
concept\_class\_id & str & The concept\_class\_id field from the OMOP concept table \\
\midrule
standard\_concept & Optional[str] & The standard\_concept field from the OMOP concept table (optional because the field can be NULL) \\
\midrule
invalid\_reason & Optional[str] & The invalid\_reason field from the OMOP concept table (optional because the field can be NULL) \\
\midrule
ranks & Optional[Dict[str, int]] & The ranking of suggestions by any algorithm used to rank concepts \\
\midrule
scores & Optional[Dict[str,int]] & The scores of suggestions by any algorithm used to rank concepts \\
\bottomrule
\end{tabular}
\caption{Suggestion fields used by Lettuce to describe suggested OMOP concepts}
\label{tab:concept_fields}
\end{table}

% Table 2: Assistant metadata fields
\begin{table}[ht]
\centering
\begin{tabular}{l l p{6cm}}
\toprule
\textbf{Field} & \textbf{Type} & \textbf{Description} \\
\midrule
assistant & str & The name of the assistant used to suggest concepts. Defaults to "Lettuce" \\
\midrule
version & str & The version number of the assistant \\
\midrule
pipeline & Optional[str] & The pipeline used to suggest concepts \\
\midrule
info & Optional[Dict[str, any]] & Any other information for metadata \\
\bottomrule
\end{tabular}
\caption{SuggestionMetaData fields}
\label{tab:metadata_fields}
\end{table}

% Table 3: Suggestions structure
\begin{table}[ht]
\centering
\begin{tabular}{l l l}
\toprule
\textbf{Field} & \textbf{Type} & \textbf{Description} \\
\midrule
recommendations & List[Suggestion] & The suggestions made by the assistant \\
\midrule
metadata & SuggestionsMetaData & Any attached metadata about the run \\
\bottomrule
\end{tabular}
\caption{Suggestions structure}
\label{tab:suggestions_structure}
\end{table}

\subsection{Lettuce Inspector}

To evaluate the performance of the Lettuce retrieval pipelines, we developed an independent experimental framework, lettuce-inspector~\parencite{HealthInformaticsUoNLettuceinspector2025}, which was used to generate all results reported in this manuscript. The framework provides a controlled environment for executing mapping tasks, configuring retrieval pipelines, and benchmarking outputs against expected gold-standard annotations or data that has been annotated and validated by data engineering teams. Lettuce-inspector ingests curated mapping datasets, executes user-defined pipeline configurations, and computes evaluation metrics tailored to OMOP concept mapping. These include both standard retrieval metrics and OMOP-specific measures, such as whether a retrieved concept is semantically related to the target concept within the vocabulary hierarchy. All experimental outputs are recorded in machine-readable JSON format to support downstream aggregation, statistical analysis, and visualisation.

\begin{figure}[hb]
    \centering
    \includegraphics[width=1\linewidth]{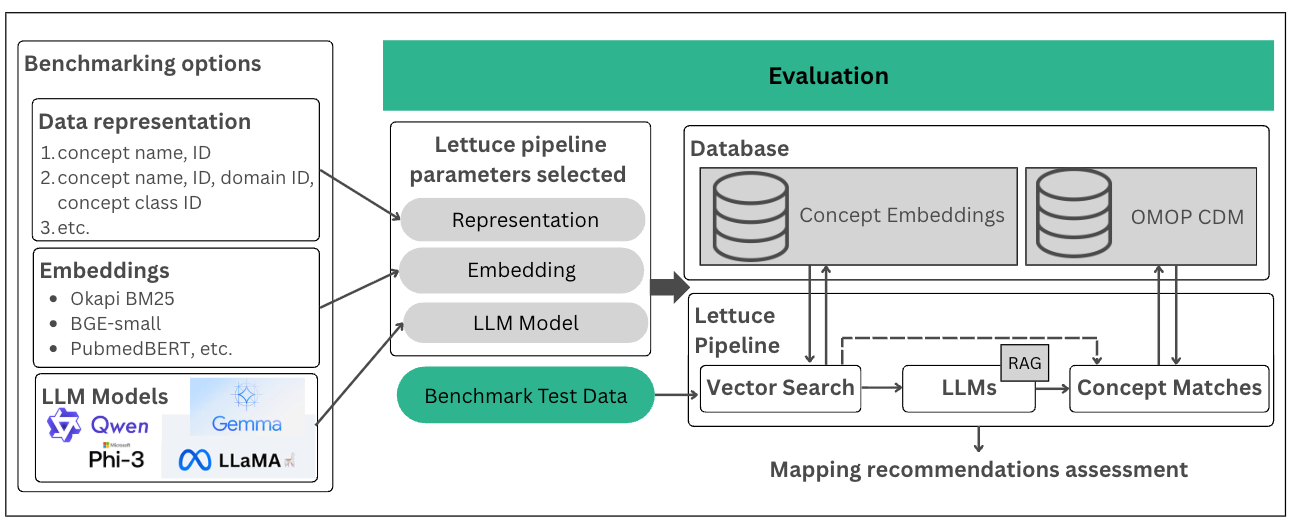}
    \caption{The Lettuce inspector framework for Lettuce benchmarking and optimisation}
    \label{fig:lettuce-inspector}
\end{figure}

Lettuce-inspector (Figure~\ref{fig:lettuce-inspector}) is designed as a modular experimentation framework that enables systematic exploration of the design and algorithmic space underlying the four retrieval pipelines described in Section~\ref{fig:lettuce-inspector}. The aim is to identify the most effective combination of input data representation, components—embedding strategy, vector search configuration, LLM model, and supporting modules---for different classes of mapping problems. The framework allows controlled investigation of trade-offs between accuracy, computational efficiency, model size, and encoding choices that influence results.

The framework enables consistent and repeatable experimentation, allowing us to quantify the effectiveness of different pipeline configurations and to determine optimised setups for specific mapping tasks while controlling for resource consumption, latency, and model complexity.

\subsection{Experimental Design}

\subsubsection{Lettuce Optimisation}

In this study, to optimise Lettuce pipelines, we evaluated several encoding configurations.
To test different kinds of retrieval, we compared lexical and semantic search.
Okapi best match 25 (BM25) ~\parencite{harmanOverviewThirdText1995} is a commonly used algorithm for text search, and is the default option for Lucene, the Java library used by Usagi, so is a reasonable representative for lexical search methods.
We compared this with embedding strategies that used as input: (i) concept identifiers and concept names, and (ii) concept names augmented with concept identifiers, domain identifiers, and concept class identifiers.
Embeddings models encode the full context of their input, so providing context in the form of the concept information may improve retrieval.
Retrieval methods were compared by their retrieval of the human-mapped concept in the top 10 suggestions.
BM25 was compared with the embedding models BGE-small, and PubMedBERT.

For the LLM-based pipeline, we evaluated the performance of several transformer models commonly used in resource-constrained environments: Qwen, Phi-3, Llama, and Gemma.
The accuracy of LLM suggestions was compared between pipelines with and without RAG from 10 concepts retrieved with each retrieval method.
All experiments were executed on two case studies to assess performance across distinct datasets.

\subsubsection{Case studies}
The case studies selected contain real world data on medications that need to be mapped into OMOP. The first dataset was obtained from Nottingham University Hospitals, in the UK; the second data was obtained from the Health for Life in Singapore.

\subsubsubsection{Nottingham University Hospitals}
A dataset of 250 drugs administered as parts of trials carried out on patients in Nottingham University Hospitals was mapped using Lettuce.
Of the original 250, 32 drugs have no matching OMOP concept, so the remaining 218 were used to test retrieval methods.

\subsubsubsection{HELIOS study}
Medication data were obtained from the Health for Life in Singapore (HELIOS) study (IRB approval by Nanyang Technological University: IRB-2016-11-030), a phenotyped longitudinal population cohort study comprising 10,004 multi-ethnic Asian population of Singapore aged 30-85 years \parencite{wangHealthLifeSingapore2025}.
Participants in the HELIOS study were recruited from the Singapore general population between 2018 and 2022 and underwent extensive clinical, behavioural, molecular and genetic characterisation.
With rich baseline data and long-term follow-up through linkage to national health data, the HELIOS study provides a unique and world class resource for biomedical researchers across a wide range of disciplines to understand the aetiology and pathogenesis of diverse disease outcomes in Asia, with potential to improve health and advance healthcare for Asian populations.

To facilitate scalable and collaborative research, the HELIOS study implements the OMOP-CDM.
However, mapping medication data to OMOP concepts poses significant challenges, primarily due to the complexities involved in standardising medication names.
In the HELIOS study, medication data were self-reported and manually entered via nurse-administered questionnaires, therefore, medications with brand name, abbreviations, typographic misspellings or phonetic errors, or combined medications could be recorded.
All of these sources of imprecision make mapping to a controlled medical vocabulary more difficult and require significant manual data cleaning.

\subsubsection{Evaluation methods}
\subsubsubsection{Retrieval}
Semantic search in Lettuce follows a similar paradigm to previous lexical search methods, like Usagi, where the best $n$ matches to a source term, as measured by some scoring function, are presented to the user.
Usagi is based on Java's Lucene library, and uses the default retrieval algorithm for Lucene, BM25~\parencite{harmanOverviewThirdText1995, robertsonSimpleBM25Extension2004}.
Two semantic search methods were compared with BM25.
From the Beijing Academy of Artificial Intelligence (BAAI), the BAAI general embedding (BGE-small) model is a small (33.4M parameter) encoder model developed for retrieval tasks~\parencite{bge_embedding}.
Using BGE-small, embeddings for each concept were created using each concept's name.
A possible weakness of this approach is that important context for each concept is held in other fields of the concept table.
For example, there are multiple concepts with the concept name ``Monster'', which refer to outdated medical terminology and to a village in the Netherlands, which this method cannot distinguish.
To overcome this limitation, a second method of generating strings for embedding was devised, using the domain and concept class from the concept table, so for each concept, the string was "\texttt{<concept name>, [a/an] <concept class id> <domain id>}".
BGE-small is a model for general use, but models trained specifically on biological and medical data might produce more relevant embeddings for OMOP mapping.
One of these, PubmedBERT-embeddings~\parencite{pubmedbert}, was trained on PubMed abstracts, and was used to generate embeddings from concept names.
Each retrieval method was assessed by how well it could return the target concept in the top 10 results.

\subsubsubsection{Model choice}
The weights for many models are available for use.
When Lettuce development started, the Llama 3.1 model with 8 billion parameters was selected.
Four models were selected for comparison by their performance on the MMLU-Pro~\parencite{wangMMLUProMoreRobust2024} benchmark: Phi-4, with 14 billion parameters; Gemma-3, with 12 billion parameters; and Qwen2.5, with 14 billion parameters~\parencite{dubeyLlamaHerdModels2024,gemmateam2025gemma3technicalreport}.
A 4-bit quantised version of each model was used.
A smaller model, Llama 3.2, with 3 billion parameters, was used for comparison, quantised to 6 bits.

\subsubsubsection{Retrieval augmented generation} \label{section:RAG}
RAG has proved useful in tasks where context is important and the whole of the potentially required context will not fit in a single prompt~\parencite{lewisRetrievalAugmentedGenerationKnowledgeIntensive2021}.
Its applicability to suggesting the correct concept from a vocabulary of millions of concepts was tested by comparing the performance of an LLM alone with that LLM with the top 10 results from semantic search inserted into the prompt.

Consistent formats were used to prompt the models in testing, using a system prompt and a user prompt.
The system prompt contains the instructions for the model.
The prompt used in these cases uses techniques recommended for Llama models~\parencite{MetallamaLlamarecipes2024}.
This part of the prompt also gives the LLM a role, which has been shown to improve consistency in responses~\parencite{kongBetterZeroShotReasoning2024}.
Importantly, the prompt includes providing examples of informal name/formal name pairs, an effective tactic for LLM prompting~\parencite{brownLanguageModelsAre2020}.

\begin{lstlisting}
   You are an assistant that suggests formal RxNorm names for a medication.
   You will be given the name of a medication<@\textcolor{red}{, along with some possibly related RxNorm terms. If you do not think these terms are related, ignore them when making your suggestion}@>.

    Respond only with the formal name of the medication, without any extra explanation.

    Examples:
    Informal name: Tylenol
    Response: Acetaminophen

    Informal name: Advil
    Response: Ibuprofen

    Informal name: Motrin
    Response: Ibuprofen

    Informal name: Aleve
    Response: Naproxen
\end{lstlisting}

The RAG prompt includes the instruction (shown in red) to attend to the additional context so that the model can apply this effectively.
Three RAG prompts were used that supplied varying context regarding the retrieved items.
The simplest supplies only the concept\_name for each concept, the second also provides the score returned by the retrieval method, and the last provides the concept's domain, vocabulary, concept class, and the score from the retrieval method as comma-separated values (CSV).
For example, for the concept \texttt{TRAZODONE HCL ORAL TABLET}, with a score of 0.051, the representations would be:

\begin{itemize}
    \item \texttt{TRAZODONE HCL ORAL TABLET}
    \item \texttt{TRAZODONE HCL ORAL TABLET (score = 0.051)}
    \item \texttt{TRAZODONE HCL ORAL TABLET, Drug, DPD, Clinical Drug Form, 0.051}
\end{itemize}

In the CSV representation, the items are preceded by a header describing the number of concepts and the column headings. The prompt follows the RAG representation (shown in red below) with the source term.

\begin{comment}
    This header overflowed the page, so I'm not describing it
\end{comment}

\begin{lstlisting}
    <@\textcolor{red}{Possible related terms:}@>
    <@\textcolor{red}{<For each search result>}@>
        <@\textcolor{red}{<RAG representation>}@>

    Informal name: {{informal_name}}
\end{lstlisting}

Each combination of LLM and each retrieval method, or the LLM alone, was evaluated for both datasets.
For the embeddings models, this included both encoding the concept\_name and the longer string with a description of the concept class and domain.
Where RAG was used, the different RAG representations were evaluated.

\subsubsubsection{Statistical analysis}
The evaluation results for all concept suggestion pipelines were used to assess whether RAG is a useful strategy for suggesting concepts.
The probability of answering with the correct concept name was modelled using binomial regression using a mixed-effects model, with the dataset the source term came from, the LLM used, and whether any RAG was used as fixed effects, and the source term as a random effect.

To assess whether the representation of RAG results affected pipeline performance, a similar model was fitted to the results of pipelines using RAG, with the dataset, LLM used, retrieval method, and RAG representation as fixed effects, and the source term as a random effect.

\section{Results}
\subsection{Retrieval}
As shown in Table \ref{tab:retrieval-accuracy-table}, semantic search outperforms lexical search for both case studies.
For the HELIOS study dataset, BM25 retrieved the target concept in the top 10 results in 10.25\% of cases, while BGE-small embeddings built on the concept name alone retrieved the target concept in 22.25\% of cases.
Introducing context to the concept by adding the concept class and domain severely impacted performance, reducing the retrieval of the target concept to 17.25\%.

For the NUH dataset, retrieval was generally more successful, as the source terms are more similar to OMOP concepts. BM25 was least successful, at 50.9\%, and BGE-small using the longer strings for embeddings most successful, at 79.8\%.

\begin{table}[h]
    \centering
    \begin{tabular}{llcc}
        \toprule
         & & \multicolumn{2}{c}{Accuracy@10} \\
         \cmidrule{3-4}
         Method & Embedding representation & HELIOS study & NUH \\
         \midrule
         BM25 & {--} & 41 (10.25\%) & 111 (50.9\%) \\
         \midrule
         \multirow{2}{*}{BGE-small} & concept name & \textbf{89} (22.25\%) & 163 (74.8\%) \\
         & with context & 69 (17.25\%) & \textbf{174} (79.8\%) \\
         \midrule
         \multirow{2}{*}{PubmedBERT} & concept name & 82 (20.5\%) & 165 (75.7\%)  \\
         & with context & 34 (8.5\%) & 115 (52.8\%) \\
         \bottomrule
    \end{tabular}
    \caption{Accuracy of retrieving the target concept in the top 10. Figures show the percentage of source terms for which the target concept was retrieved in the top 10 results for each retrieval method.}
    \label{tab:retrieval-accuracy-table}
\end{table}

Scores were much higher for all methods running on the NUH dataset.
BM25 retrieves the target concept in 50.9\% of cases, and the best scoring semantic search, BGE-small with context, scores 79.8\%.

\subsection{Model choice}
Models fitting different resource constraints were selected, and Table \ref{tab:accuracy-table} shows the performance of the different models.
Greater model size did not correlate with better accuracy on the HELIOS study dataset.
In fact, the worst performance for the HELIOS study dataset came from the Phi-4 model, which scored 26.5\% without RAG.
Gemma has the best performance without RAG on this dataset, at 35\%.

\begin{longtable}{
    l 
    l 
    p{3.8cm} 
    S[table-format=2.2, detect-weight, mode=text] 
    S[table-format=2.2, detect-weight, mode=text]
    }

    \toprule
    & & & \multicolumn{2}{c}{Accuracy (\%)} \\
    \cmidrule(lr){4-5}
    {Retrieval} & {Embedding} & {RAG Representation} & {HELIOS study} & {NUH} \\
    \midrule
    \endfirsthead

    \toprule
    & & & \multicolumn{2}{c}{Accuracy (\%)} \\
    \cmidrule(lr){4-5}
    {Retrieval} & {Embedding} & {RAG Representation} & {HELIOS study} & {NUH} \\
    \midrule
    \endhead

    \bottomrule
    \endfoot

    \caption{Percentage accuracy of LLMs generating the target concept. Best results per dataset in \textbf{bold}, best method per model in \textit{italics}.}
    \label{tab:accuracy-table}
    \endlastfoot

    % ================= Llama 3.1 8b =================
    \multicolumn{5}{l}{\textbf{Llama 3.1 8b}} \\
    \midrule
    \multicolumn{3}{l}{\textit{Base Model (No Retrieval)}} & 30 & 55.96 \\
    \addlinespace[0.5em]
    
    BM25 & {--} & Concept names & 37 & 85.78 \\
         &      & Names and scores & \textit{40.5} & 85.32 \\
         &      & Concept attributes & 38 & 86.7 \\
    \addlinespace
    
    BGE-small & Concept name & Concept names & 35.75 & 87.16 \\
              &              & Names and scores & 38.25 & 85.32 \\
              &              & Concept attributes & 38 & \textit{88.53} \\
    \addlinespace[0.2em]  
              & With context & Concept names & 36.25 & 86.7 \\
              &              & Names and scores & 35.25 & 87.16 \\
              &              & Concept attributes & 38.25 & 86.7 \\
    \addlinespace
    
    PubmedBERT & Concept name & Concept names & 37.25 & 87.16 \\
               &              & Names and scores & 39.25 & 87.16 \\
               &              & Concept attributes & 40 & 88.07 \\
    \addlinespace[0.2em]   
               & With context & Concept names & 33.25 & 81.19 \\
               &              & Names and scores & 34.25 & 80.73 \\
               &              & Concept attributes & 37.75 & 83.94 \\
    
    \midrule
    
    % ================= Llama 3.2 3b =================
    \multicolumn{5}{l}{\textbf{Llama 3.2 3b}} \\
    \midrule
    \multicolumn{3}{l}{\textit{Base Model (No Retrieval)}} & 28.75 & 57.34 \\
    \addlinespace[0.5em]

    BM25 & {--} & Concept names & 35.5 & 72.94 \\
         &      & Names and scores & 35 & 75.69 \\
         &      & Concept attributes & 35 & 76.61 \\
    \addlinespace

    BGE-small & Concept name & Concept names & 33.75 & 74.31 \\
              &              & Names and scores & 33.75 & 77.52 \\
              &              & Concept attributes & 36 & \textit{81.19} \\
    \addlinespace[0.2em]
              & With context & Concept names & 34.25 & 77.06 \\
              &              & Names and scores & \textit{36.75} & 68.35 \\
              &              & Concept attributes & 35.5 & 79.82 \\
    \addlinespace
    
    PubmedBERT & Concept name & Concept names & 36 & 75.23 \\
               &              & Names and scores & 35.25 & 75.23 \\
               &              & Concept attributes & 34.5 & 77.52 \\
    \addlinespace[0.2em]
               & With context & Concept names & 32 & 70.18 \\
               &              & Names and scores & 32.75 & 69.27 \\
               &              & Concept attributes & \textit{36.75} & 79.82 \\

    \midrule

    % ================= Gemma 3 12b =================
    \multicolumn{5}{l}{\textbf{Gemma 3 12b}} \\
    \midrule
    \multicolumn{3}{l}{\textit{Base Model (No Retrieval)}} & 35 & 64.22 \\
    \addlinespace[0.5em]
    
    BM25 & {--} & Concept names & 44 & 87.61 \\
         &      & Names and scores & 43.25 & 87.61 \\
         &      & Concept attributes & \textbf{45.5} & 88.07 \\
    \addlinespace
    
    BGE-small & Concept name & Concept names & 38.75 & 87.61 \\
              &              & Names and scores & 39.75 & 86.24 \\
              &              & Concept attributes & 39.25 & 88.99 \\
    \addlinespace[0.2em]
              & With context & Concept names & 40.75 & \textbf{89.91} \\
              &              & Names and scores & 42.75 & \textbf{89.91} \\
              &              & Concept attributes & 39.75 & \textbf{89.91} \\
    \addlinespace
    
    PubmedBERT & Concept name & Concept names & 41.75 & 88.07 \\
               &              & Names and scores & 39.75 & 87.16 \\
               &              & Concept attributes & 42 & 88.07 \\
    \addlinespace[0.2em]
               & With context & Concept names & 40.5 & 84.86 \\
               &              & Names and scores & 41 & 84.4 \\
               &              & Concept attributes & 42.25 & 86.24 \\
    
    \midrule

    % ================= Phi-4 =================
    \multicolumn{5}{l}{\textbf{Phi-4}} \\
    \midrule
    \multicolumn{3}{l}{\textit{Base Model (No Retrieval)}} & 26.5 & 59.17 \\
    \addlinespace[0.5em]

    BM25 & {--} & Concept names & 27.75 & 72.48 \\
         &      & Names and scores & 28 & 74.77 \\
         &      & Concept attributes & 30 & 73.39 \\
    \addlinespace

    BGE-small & Concept name & Concept names & 28 & 71.1 \\
              &              & Names and scores & 27.5 & 72.48 \\
              &              & Concept attributes & 30.75 & 75.23 \\
    \addlinespace[0.2em]
              & With context & Concept names & 28.25 & 72.48 \\
              &              & Names and scores & \textit{31.75} & 73.85 \\
              &              & Concept attributes & 31.25 & 73.39 \\
    \addlinespace
    
    PubmedBERT & Concept name & Concept names & 28.75 & \textit{76.15} \\
               &              & Names and scores & 29.25 & 74.77 \\
               &              & Concept attributes & 31.25 & \textit{76.15} \\
    \addlinespace[0.2em]
               & With context & Concept names & 25.75 & 71.56 \\
               &              & Names and scores & 29.5 & 68.35 \\
               &              & Concept attributes & 29.25 & 71.56 \\

    \midrule

    % ================= Qwen2.5 14b =================
    \multicolumn{5}{l}{\textbf{Qwen2.5 14b}} \\
    \midrule
    \multicolumn{3}{l}{\textit{Base Model (No Retrieval)}} & 30.5 & 75.23 \\
    \addlinespace[0.5em]

    BM25 & {--} & Concept names & 32.5 & 83.03 \\
         &      & Names and scores & 30.75 & 84.4 \\
         &      & Concept attributes & \textit{34.25} & 84.4 \\
    \addlinespace
    
    BGE-small & Concept name & Concept names & 29 & 80.73 \\
              &              & Names and scores & 32.5 & 81.19 \\
              &              & Concept attributes & 29.75 & 80.73 \\
    \addlinespace[0.2em]
              & With context & Concept names & 30.25 & 83.03 \\
              &              & Names and scores & 30.75 & \textit{84.86} \\
              &              & Concept attributes & 31.25 & 83.03 \\
    \addlinespace
    
    PubmedBERT & Concept name & Concept names & 31.75 & 82.11 \\
               &              & Names and scores & 33 & 83.94 \\
               &              & Concept attributes & 33.5 & \textit{84.86} \\
    \addlinespace[0.2em]
               & With context & Concept names & 30.25 & 74.31 \\
               &              & Names and scores & 30.25 & 77.06 \\
               &              & Concept attributes & 30.5 & 77.52 \\

    \bottomrule
\end{longtable}

\subsection{Retrieval augmented generation}
The best performance was achieved by Gemma using RAG.
For the HELIOS study dataset, using BM25 as a retrieval method, and supplying the concept attributes, was most successful at 45.5\%.
This was also the best RAG method for Qwen2.5, which achieved 30.5\% accuracy with no RAG, and 34.25\% with this combination.
On the NUH dataset, using the BGE-small retriever with context in the embeddings was most successful, but the representation of concepts in the prompt did not affect the outcome.

Across the datasets, the odds ratio for a correct answer with RAG compared with without was 1.62 (\textit{P}<.001), meaning the odds of the LLM providing the correct answer were increased by 62\% when RAG is enabled compared with the same LLM without RAG.
Compared with supplying only the names of retrieved concepts in the prompt, supplying the score has an odds ratio of 1.03 (\textit{P}=.008), and supplying concept attributes has an odds ratio of 1.09 (\textit{P}<.001).

%--- Section ---%
\newpage
\section{Discussion}
Lettuce pipelines perform well on different kinds of datasets in the drug domain.
The NUH dataset shows that source terms taken from a formal vocabulary can be matched well using existing lexical search methods, with a moderate increase in performance using semantic search.
By contrast, the performance on the HELIOS study dataset shows that lexical search is inadequate to the task of matching idiosyncratic, informal source terms, whereas semantic search can find the correct concept in the top 10 results around twice as often, and an LLM-enabled pipeline can suggest the correct concept in almost half of cases.
New, more capable models are constantly released, and different users will have different resource requirements, so making specific recommendations to use a particular LLM is futile.
We have presented the lettuce-inspector evaluations to demonstrate how Lettuce configurations can be tested.
The results do show that providing the LLM with possible concepts using RAG is a generally successful strategy, controlling for the model used and dataset.
Surprisingly, a retrieval method's performance was not a reliable indicator of how well it would perform in RAG; BM25 had the lowest accuracy of all retrieval methods on the HELIOS study dataset, half that of BGE-small, but was not significantly worse than other retrieval methods for RAG.
Configuring the representation of source terms provides more marginal benefits.
Adding the concept scores increases the odds of a correct response with a very modest effect, suggesting the LLM can use the scores as a slight hint, but that simply providing the names provides most of the benefit.
Adding in concept attributes is more helpful, suggesting the models can leverage the added context to make better outputs.

As Lettuce development continues, it will be important to design pipelines that are flexible to different OMOP domains.
For example, in the retrieval tests, using a longer string to reduce retrieval of concepts with similar concept names from different domains could severely decrease performance, but in domains with less discrete concepts than drugs, this may be necessary.
Alternatively, ancillary methods may be employed, such as hierarchical classifiers~\parencite{sillaSurveyHierarchicalClassification2011} which estimate the domain and concept class of a source term, which can then be used to narrow down vector search.
These classifiers provide an example of a strength of Lettuce's modular, trainable approach.
As well as testing specialist models for biomedical terms, such as PubmedBERT, as we have done for the drug domain, Lettuce can employ models fine-tuned on OMOP specific tasks, using lettuce user feedback for training.

In production, the prompt for Lettuce does not specify a target vocabulary, instead using a dynamic prompt that specifies a domain or domains if specified in the request, or ``source term'' if no domain is specified.
Prompt engineering can dramatically increase performance of LLM pipelines, and finding a prompt that can effectively instruct an LLM to suggest the correct concept across domains will require a systematic approach.
This prompt may include the results of classifiers to help the LLM narrow down domains.
As more tools are added to the context provided to the LLM, it may make more sense to implement Lettuce as an agent which can employ these tools at its own discretion.
Agentic LLM pipelines provide a flexible solution to complex problems, and using methods like the Model Context Protocol (MCP), new tools can easily be integrated.
With automation and semi-automation in data pipelines, knowing the reasons for decisions made becomes increasingly important for transparency and reproducibility.
LLMs cannot reliably report on their own reasoning~\parencite{chenReasoningModelsDont2025}, presenting a data provenance problem for pipelines involving LLMs.
However, tracking an agent's tool usage and recording the requests and results will provide a self-annotating decision process, including any human decisions made in mapping, which could inherently provide more insight into the reasons for mapping than a manual pipeline.

These future versions of Lettuce will be developed with careful consideration of the needs of the environments in which they need to be deployed.
The Lettuce API has been designed to integrate with Carrot-mapper, and this integration will continue to drive the format of the API.
%-------------------------------------------
% Optional Contents
%-------------------------------------------
%--- Section ---%
\section*{Funding}
This research was funded by the National Institute for Health and Care Research Nottingham Biomedical Research Centre and Health Data Research UK (HDR-23002), which is funded by the Medical Research Council (UKRI), the National Institute for Health Research, the British Heart Foundation, Cancer Research UK, the Economic and Social Research Council (UKRI), the Engineering and Physical Sciences Research Council (UKRI), Health and Care Research Wales, Chief Scientist Office of the Scottish Government Health and Social Care Directorates, and Health and Social Care Research and Development Division (Public Health Agency, Northern Ireland). .

\section*{Conflicts of Interest}
None declared

%--- Section ---%
\section*{Data Availability}
Data access requests can be submitted to the HELIOS Data Access Committee by emailing helios\_science@ntu.edu.sg for details. 

\subsection*{Code Availability}
The code used to generate the results described can be found in the lettuce-inspector repository~\parencite{HealthInformaticsUoNLettuceinspector2025}.

Code for the statistical analysis performed can be found at doi:10.5281/zenodo.18759519

%--- Section ---%
\section*{Authors' contributions}

\begin{itemize}
%Conceptualization – Ideas, formulation or evolution of overarching research goals and aims.
    \item Conceptualization: GF, PRQ, JMW, EU

%Data curation – Management activities to annotate (produce metadata), scrub data and maintain research data (including software code, where it is necessary for interpreting the data itself) for initial use and later re-use.
    \item Data curation: JMW

%Formal analysis – Application of statistical, mathematical, computational, or other formal techniques to analyze or synthesize study data.
    \item Formal analysis: JMW, GF, DGG

%Funding acquisition ​- Acquisition of the financial support for the project leading to this publication.
    \item Funding acquisition: PRQ, GF

%Investigation – ​Conducting a research and investigation process, specifically performing the experiments, or data/evidence collection.
    \item Investigation: JMW

%Methodology – Development or design of methodology; creation of models.
    \item Methodology: GF, JMW

%Project administration – Management and coordination responsibility for the research activity planning and execution.
    \item Project administration: GF

%Resources – Provision of study materials, reagents, materials, patients, laboratory samples, animals, instrumentation, computing resources, or other analysis tools.

%would this be the HELIOS people?
    \item Resources: AR, RL, XW, TM, JC

%Software – Programming, software development; designing computer programs; implementation of the computer code and supporting algorithms; testing of existing code components.
    \item Software: JMW, RO, BP, KS, AR, TG

%Supervision – Oversight and leadership responsibility for the research activity planning and execution, including mentorship external to the core team.
    \item Supervision: PRQ, GF

%Validation – Verification, whether as a part of the activity or separate, of the overall replication/reproducibility of results/experiments and other research outputs.
%
%Visualization – Preparation, creation and/or presentation of the published work, specifically visualization/data presentation.
    \item Visualization: GF, JMW, DGG

%Writing – original draft – ​Preparation, creation and/or presentation of the published work, specifically writing the initial draft (including substantive translation).
    \item Writing - original draft: GF, JMW, EU, PRQ

%Writing - review & editing – Preparation, creation and/or presentation of the published work by those from the original research group, specifically critical review, commentary or revision – including pre- or post-publication stages.
    \item Writing - review \& editing: All authors contributed to review and editing of the manuscript

\end{itemize}
%--- Section ---%
\section*{Abbreviations}

\begin{itemize}
    \item \textbf{API:} Application programming interface
    \item \textbf{BM25:} Okapi Best match 25
    \item \textbf{CDM:} Common data model
    \item \textbf{CLI:} Command-line interface
    \item \textbf{EHR:} Electronic health record
    \item \textbf{FAIR:} Findable, accessible, interoperable, reusable
    \item \textbf{HELIOS:} Health for Life in Singapore (study)
    \item \textbf{LLM:} Large language model
    \item \textbf{NUH:} Nottingham University Hospitals
    \item \textbf{OHDSI:} Observational Health Data Sciences and Informatics
    \item \textbf{OMOP:} Observable Medical Outcomes Partnership
    \item \textbf{RAG:} Retrieval augmented generation
\end{itemize}

%-------------------------------------------
% References
%-------------------------------------------

% Print bibliography
\printbibliography

%-------------------------------------------
% Appendix
%-------------------------------------------
% Activate the appendix in the doc
% from here on sections are numerated with capital letters 
%\appendix

% Change equation numbering format to be sequential within sections in the appendix
\renewcommand\theequation{\Alph{section}\arabic{equation}} % Redefine equation numbering format
\counterwithin*{equation}{section} % Number equations within sections
\renewcommand\thefigure{\Alph{section}\arabic{figure}} % Redefine equation numbering format
\counterwithin*{figure}{section} % Number equations within sections
\renewcommand\thetable{\Alph{section}\arabic{table}} % Redefine equation numbering format
\counterwithin*{table}{section} % Number equations within sections

\end{document}